# Reinforcement Learning From State and Temporal Differences *

**Lex Weaver**
Department of Computer Science
Australian National University
ACT AUSTRALIA 0200
*Lex.Weaver@anu.edu.au*

**Jonathan Baxter**
Computer Sciences Laboratory
Australian National University
ACT AUSTRALIA 0200
*Jonathan.Baxter@anu.edu.au*

**Abstract**

TD($\lambda$) with function approximation has proved empirically successful for some complex reinforcement learning problems. For linear approximation, TD($\lambda$) has been shown to minimise the squared error between the approximate value of each state and the true value. However, as far as policy is concerned, it is error in the relative ordering of states that is critical, rather than error in the state values. We illustrate this point, both in simple two-state and three-state systems in which TD($\lambda$)—starting from an optimal policy—converges to a sub-optimal policy, and also in backgammon. We then present a modified form of TD($\lambda$), called STD($\lambda$), in which function approximators are trained with respect to relative state values on binary decision problems. A theoretical analysis, including a proof of monotonic policy improvement for STD($\lambda$) in the context of the two-state system, is presented, along with a comparison with Bertsekas' *differential training* method [1]. This is followed by successful demonstrations of STD($\lambda$) on the two-state system and a variation on the well known acrobot problem.

## 1 Introduction

For complex reinforcement learning problems, TD($\lambda$) with function approximation [2] has proved empirically successful. Its origins go back as far as Samuel's Checkers Program [3], while perhaps its most famous success has been Tesauro's TD-Gammon [4, 5]. A variant of TD($\lambda$) for minimax search has also been successful in learning to play chess [6]. Successes outside of the games domain include job-shop scheduling [7] and dynamic channel allocation [8].

For linear approximation, TD($\lambda$) has been shown to minimise the squared error between the approximate value of each state and the true value [9, 10, 11, 12]. However, as far as policy is concerned, it is error in the relative ordering of states that is critical, rather than error in the state values. Consider a simple system consisting of two-states: $A$ and $B$, where the value of each state under the optimal policy is 10 and 5 respectively. A function approximator which estimated the state values to be 15 and 0, would implement the optimal policy whilst having a squared error of $5^2$. However, a function approximator estimating the values as 7 and 8, would have a squared error of only $3^2$, yet would not implement the optimal policy.

We illustrate this point further in section 2 with simple two-state and three-state systems in which TD($\lambda$), starting from the optimal policy, converges to a sub-optimal policy. In section 3 we demonstrate that this also occurs in a more complex system: backgammon. In section 4 we present a modified form of TD($\lambda$), called State-Temporal Difference($\lambda$) (or simply STD($\lambda$)), in which function approximators are trained with respect to relative state values on binary decision problems. We present results characterising the limiting behaviour of STD($\lambda$) and provide a proof of monotonic policy improvement for the two-state system of section 2. Experimental results in section 5 demonstrate the success of STD($\lambda$) in the two-state system and in a variant of the well known acrobot problem.

## 1.1 Previous work on State Differences.

Several researchers have previously considered using state difference information. Utgoff and Saxena [13] described the Best-First Preference Learning Algorithm for creating a set of preference predicates from a previously known solution. Utgoff and Clouse [14] showed how a similar algorithm can be used to generate weight updates for a linear function approximator. The technique relies upon an external expert to nominate the preferred states, and thus is a form of supervised learning. They have also constructed an algorithm which has a supervised learning phase to utilise expert preferences, and a reinforcement learning phase to make traditional TD-style updates which don't use the expert preferences. Tesauro [15] described an experiment in which he used expert preferences to train a neural network to choose between backgammon positions.

The approximation of cost-to-go differences—rather than just the cost-to-go—has been investigated by a number of authors in the context of $Q$-learning. Werbos [16, 17] discusses the merits of approximating the gradient of the $Q$ function, whilst Baird [18], Harmon, and Klopf [19] introduced *advantaging updating* which estimates the value of each state and the relative advantage of each action using separate approximation architectures.

More recently McGovern and Moss [20] have used temporal difference learning to develop an instruction scheduler for an optimising compiler. Their approach uses table-lookup rather than function approximation, and combines possible successor states into a single feature vector which is mapped to a preference indicator.

Bertsekas' *differential training* [1] is the most closely related previous work. We defer discussion of it until section 4.3.

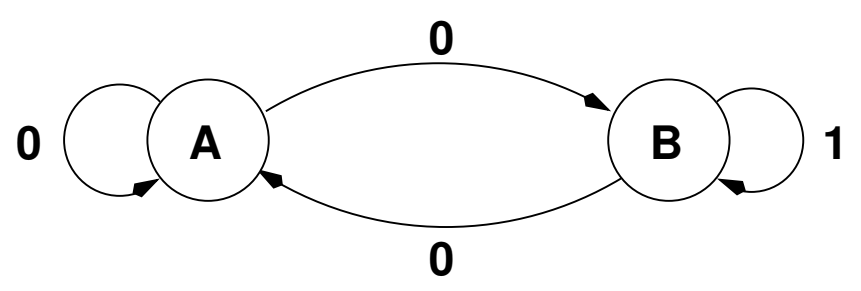

Figure 1: Transitions and rewards in the two-state system.

## 2 Generating Sub-optimal Policies

Convergence by TD($\lambda$) to sub-optimal policies can be found in even the simplest non-trivial system — a Markov Decision Process with only two states. Consider the transition matrix shown in Table 1. The corresponding machine, shown in Figure 1, has two states, $A$ and $B$, and in each state there are two actions to choose from. Action $a_1$ takes the machine to state $A$ with probability $0.8$ and action $a_2$ takes it to state $B$ with the same probability.

| Current State | Action | Destination State Probabilities $A$ | $B$ |
|---|---|---|---|
| $A$ or $B$ | $a_1$ | 0.8 | 0.2 |
| $A$ or $B$ | $a_2$ | 0.2 | 0.8 |

Table 1: Transition Matrix of the Two-state System

The rewards for each state transition are shown on the links in Figure 1 . Denoting the reward for a transition from state $x$ to state $x'$ by $r(x, x')$, we have $r(A, A) = r(A, B) = r(B, A) = 0$, and $r(B, B) = 1$. If we assume only deterministic policies, there are four possible policies for this system: always choose action $a_1$, always choose action $a_2$, choose $a_1$ in state A and $a_2$ in state B, or vice-versa.

In this paper we only consider infinite-horizon, discounted-reward problems, so the value of a state $x$ under policy $\mu$, $J^\mu(x)$, is defined to be the expected discounted reward obtained by starting in that state and following the policy $\mu$:

$$J^\mu(x) = \mathbb{E}^\mu \left[ \sum_{t=0}^{\infty} \alpha^t r(x_t, x_{t+1}) | x_0 = x \right],$$

where $x_t$ is the current state at time $t$, $\mathbb{E}^\mu$ denotes the expectation over trajectories $x_0, x_1, \ldots$ under policy $\mu$, and $\alpha \in [0, 1)$ is the discount factor.

For this simple system the optimal policy is to always choose the action leading to state B with highest probability, that is to always choose action $a_2$. Let $J^\mu(x)$ denote the value of state $x$ under the optimal policy. The following relations between $J^\mu(A)$ and $J^\mu(B)$ must hold:

$$J^\mu(A) = p(A, B)\alpha J^\mu(B) + p(A, A)\alpha J^\mu(A)$$
$$J^\mu(B) = p(B, B)(1 + \alpha J^\mu(B)) + p(B, A)\alpha J^\mu(A),$$

where $p(x, x')$ denotes the probability of a transition from state $x$ to $x'$ under the optimal policy.

Substituting for $p(\cdot)$ from Table 1 and simplifying, we can express $J^\mu(\cdot)$ in terms of $\alpha$:

$$J^\mu(A) = J^\mu(B) - 0.8 \tag{1}$$

$$J^\mu(B) = \frac{0.8 - 0.16.\alpha}{1-\alpha} \tag{2}$$

Note that $J^\mu(B) > J^\mu(A)$ for all $\alpha \in [0,1)$.

We define our approximate linear value function $\tilde{J}(x,w)$ by

$$\tilde{J}(x,w) = w \cdot \phi(x) = \sum_{i=1}^{k} w_i \phi_i(x),$$

where $w = (w_1, \ldots, w_k)$ is the parameter vector, and $\phi(x) = (\phi_1(x), \ldots, \phi_k(x))$ is a function mapping states to feature vectors. In our simple two state system we require the dimensionality of $w$ and $\phi$ to be 1 (*i.e.* $k = 1$) in order to ensure that $\tilde{J}$ cannot approximate all possible value functions (*i.e.* to ensure we really are in an *approximate* value-function setting). Hence our approximate value function is simply $\tilde{J}(x,w) = w\phi(x)$ for scalar $w$ and $\phi$. For our example we take

$$\phi(A) = 2 \quad \text{and} \quad \phi(B) = 1. \tag{3}$$

If we generate a policy from $\tilde{J}(x,w)$ by using one-step greedy look-ahead, then to prefer action $a_2$ over action $a_1$, we require $\tilde{J}(B,w) > \tilde{J}(A,w)$, which will hold only if $w < 0$. Hence, for a learning algorithm to yield an approximator which implements the optimal policy (ie: correctly values state $B$ above state $A$), it must tune $w$ to a negative value.

Assuming the optimal policy is being followed, and with $\mathrm{TD}(1)$ as the learning algorithm, [9, Theorem 1] shows that $w$ will converge to:

$$w^* = \text{argmin}_w \|\tilde{J}(\cdot,w) - J^\mu\|_D,$$

where

$$\|\tilde{J}(\cdot,w) - J^\mu\|_D := \sum_x \pi(x)(\tilde{J}(x,w) - J^\mu(x))^2,$$

where $\pi(x)$ is the stationary probability of state $x$. That is, $\mathrm{TD}(1)$ converges to a parameter vector minimising the weighted least-squared error between the approximate value of a state and its true value, where each state is weighted by its stationary probability.

For $\alpha = 0.5$, a quick calculation gives $w^* = 0.88$ — the wrong sign. Figure 2 shows the evolution of $(\tilde{J}(A,w), \tilde{J}(B,w))$ plotted after every 50 iterations of TD($\lambda$). The system converges to $(1.76, 0.88)$, minimising the squared error to $J^\mu(A) = 0.64$ and $J^\mu(B) = 1.44$ — the true values under the optimal policy.

Note that the system started in the region of optimal policy, with $w < 0$ and $(\tilde{J}(A,w), \tilde{J}(B,w))$ above the line $\tilde{J}(A,q) = \tilde{J}(B,w)$. However, TD($\lambda$) moved it into the sub-optimal policy region to minimise the squared error on the state value approximations.

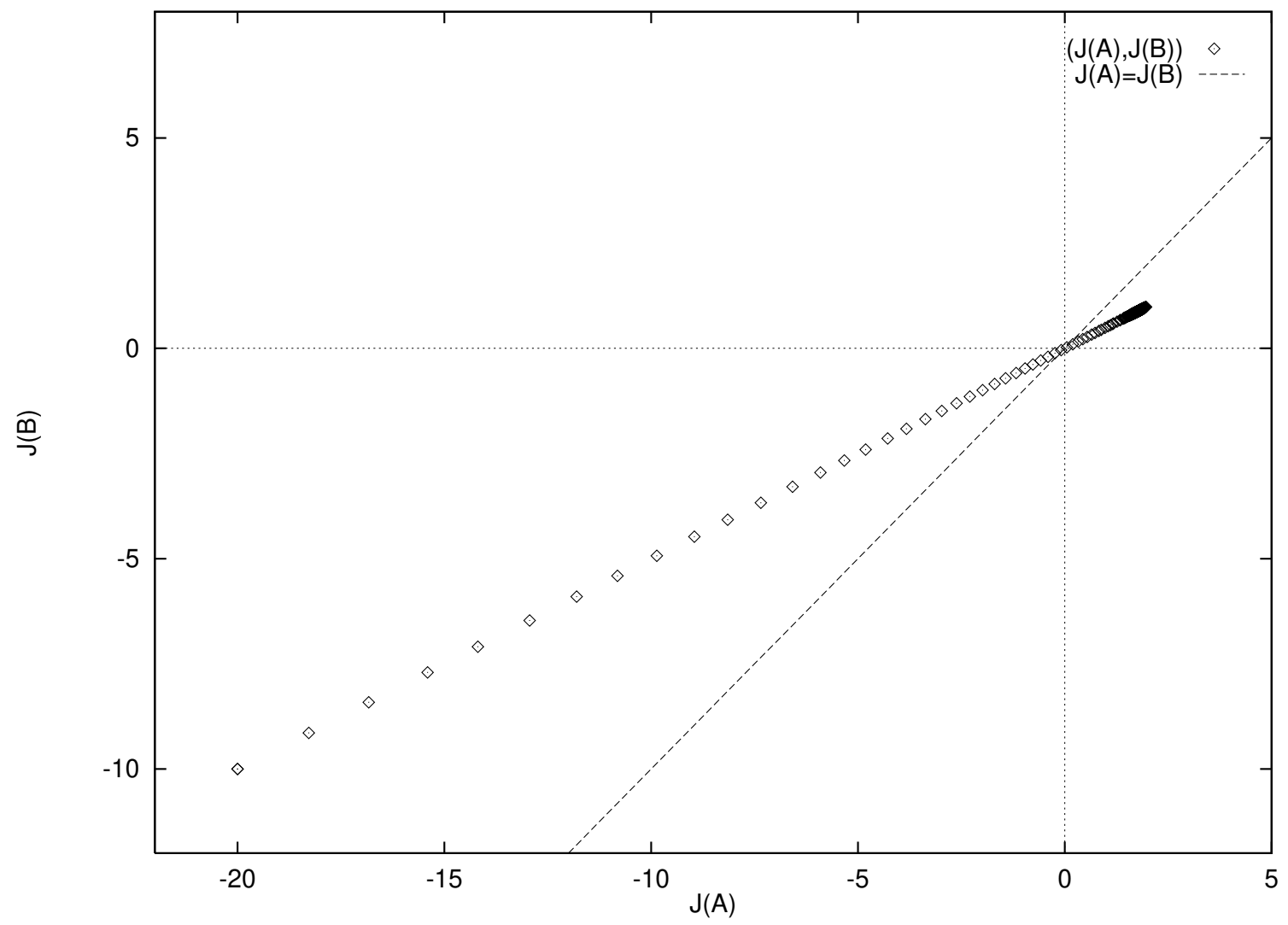

Figure 2: Evolution of $(\tilde{J}(A,w), \tilde{J}(B,w))$ using TD($\lambda$) and starting from $w = -10$, ie:$(-20, -10)$, plotted every 50 iterations. The region above $\tilde{J}(A,w) = \tilde{J}(B,w)$ gives optimal policy.

| Origin State | Action | Destination State Probabilities | | |
|---|---|---|---|---|
| | | $A$ | $B$ | $C$ |
| $A$ | $a1$ | 0.0 | 0.8 | 0.2 |
| $A$ | $a2$ | 0.0 | 0.2 | 0.8 |
| $B$ | $a1$ | 0.8 | 0.0 | 0.2 |
| $B$ | $a2$ | 0.2 | 0.0 | 0.8 |
| $C$ | $a1$ | 0.0 | 0.8 | 0.2 |
| $C$ | $a2$ | 0.0 | 0.2 | 0.8 |

Table 2: Transition Matrix of the Three-state System

## 2.1 The Three-state System

We have also constructed a three-state system in which TD($\lambda$) exhibits the same behaviour. We feel this system is interesting because it makes the learning problem subtly more difficult. In the two-state system, each observation of action choice can be interpreted as a full ordering on the desirability of the states. In the three-state system, each action choice only orders the desirability of two states. In particular, one pairing of states ($A$ and $B$) is never encountered, and a preference (if any) between these states must be inferred by the learning algorithm.

The transition matrix for the three-state system is shown in Table 2, with the transitions and rewards shown on the diagram in Figure 3.

Again, always selecting action $a2$ is the optimal policy. However, we now have two features rather than one; the linear evaluator is still of the form $J = w\Phi$, but $w$ and $\Phi$ are vectors. Hence, we have $J = w_1\phi_1(X) + w_2\phi_2(X)$.

Working from the values in Table 3, it can be confirmed that state $C$, the most

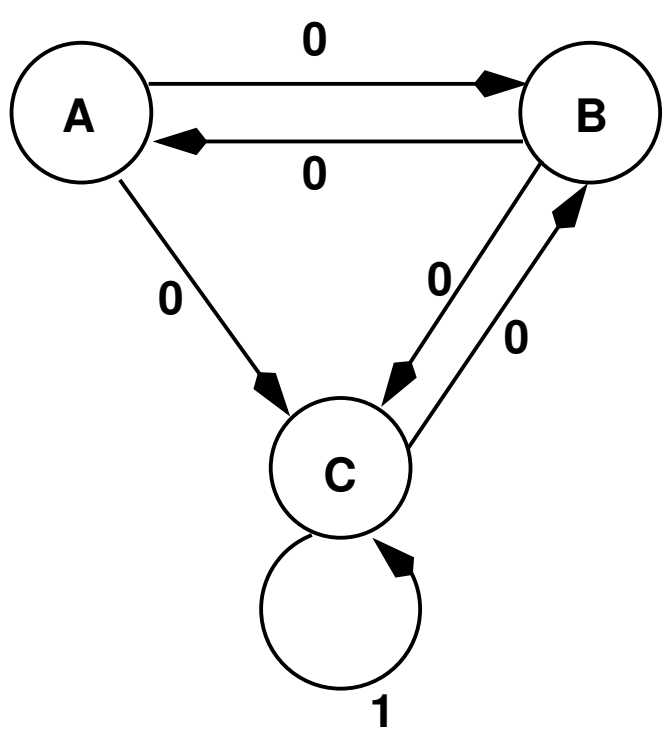

Figure 3: Transitions and rewards in the three-state system.

$$\begin{array}{ll} \phi_1(A) = \frac{12}{18} & \phi_2(A) = \frac{6}{18} \\ \phi_1(B) = \frac{6}{18} & \phi_2(B) = \frac{12}{18} \\ \phi_1(C) = \frac{5}{18} & \phi_2(C) = \frac{5}{18} \\ \alpha = 0.95 & \end{array}$$

Table 3: Three-state System values used to generate Figure 4.

likely outcome of action $a2$, is the highest valued state: $J^*(A) = J^*(B) = 12.16$ whilst $J^*(C) = 12.96$. Our choices for $\phi_1$ and $\phi_2$ however, ensure that positive values of $w_1$ and $w_2$ cannot yield a $J$ consistent with this ordering. The correct ordering can only be achieved by negative values satisfying $w_1 = w_2$.[1]

Again, tuning $w$ with TD($\lambda$) results in convergence to an approximator which implements a sub-optimal policy. Figure 4 shows a typical example of this. $w$ was initialised to (-10,-10), implementing the optimal policy even though $\|J - J^*\|_D$ wasn't minimised. $w$ was then plotted on the $w_1 w_2$ plane after every 500 iterations of TD($\lambda$), for a total of two million iterations. It converges to a point in the positive quadrant, approximately $(35.27, 4.46)$, which minimises $\|J - J^*\|_D$ but implements a sub-optimal policy.

# 3 Sub-optimal Policies in Backgammon

The example of the previous section demonstrates that it is possible for TD($\lambda$) to degrade the policy whilst still minimising $\|\tilde{J} - J^\mu\|_D$, where $J^\mu$ is the true value function of the system being observed. We now show that this behaviour affects not only artificial examples, but is also evident in a real domain: backgammon — where TD($\lambda$) has had possibly its most famous success.

Our backgammon playing program has been created along the lines of Tesauro's TD-Gammon [4, 5]. Its neural network function approximator has 209 input nodes, no hidden layer, and 1 output node, which is a squashed linear function of the inputs. The input vector consists of 200 elements directly representing the board, 8 elements representing hand-coded features extracted from the board (e.g: degree of contact, probability of hitting a blot), and a constant value 1 (from the bias).

[1]Since $J^*(A) = J^*(B)$, we could relax the ordering to the conjunction $(J(C) > J(A)) \wedge (J(C) > J(B))$ without loss. This admits regions of the $w_1 w_2$ plane on either side of the negative portion of the $w_1 = w_2$ line, but none of the positive quadrant.

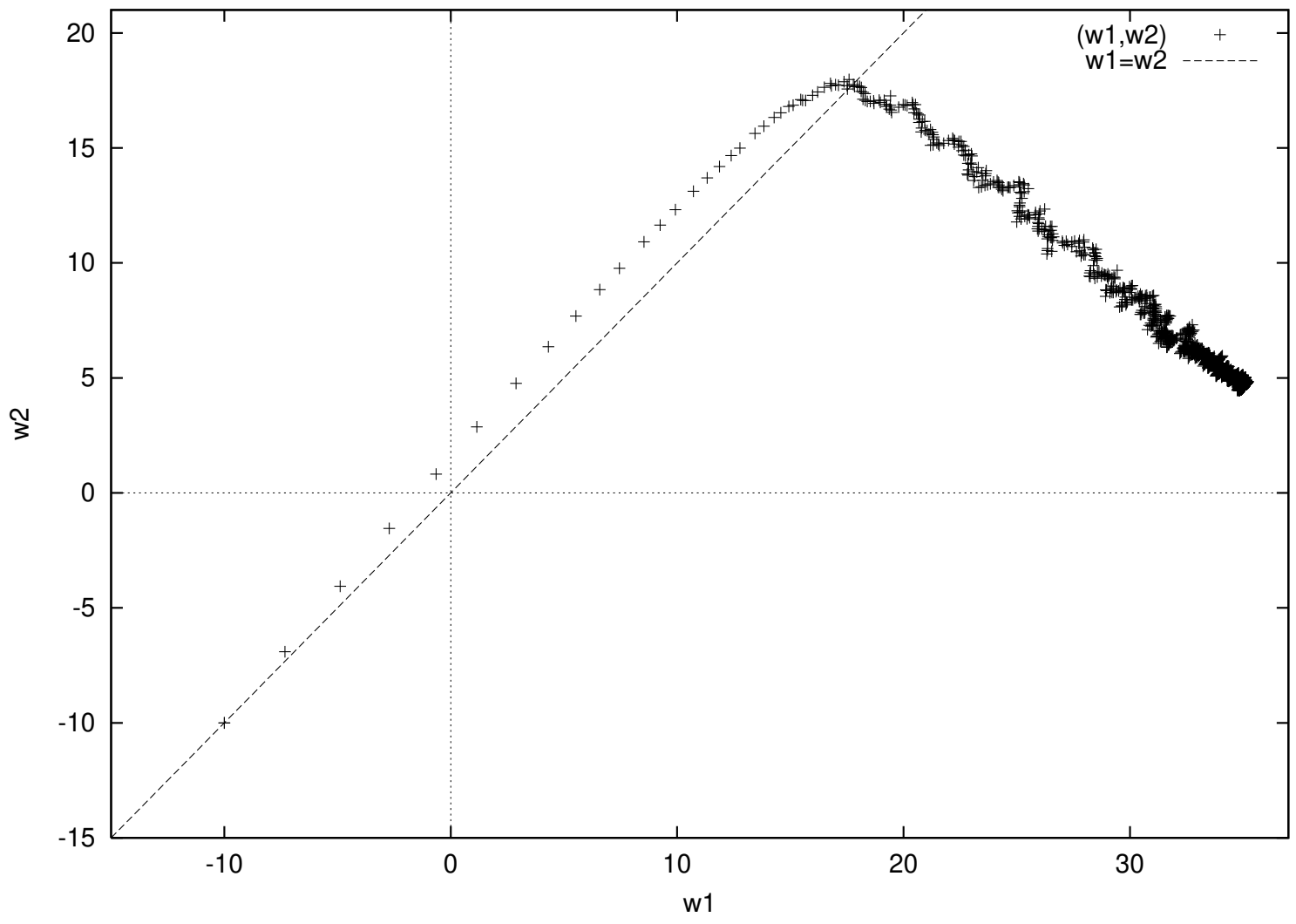

Figure 4: Evolution of $w$ using TD($\lambda$) and starting from $(-10, -10)$, plotted every 500 iterations.

We used TD($\lambda$) to train a large number of randomly generated neural networks using self-play, and monitored how their level of performance changed as training progressed. We measured performance by playing the network against a hand-coded fixed opponent (PUBEVAL, available from `http://www.revolver.demon.co.uk/bg/bot/b_pubeval.html`) for at least 2000 games, recording the proportion of games won, ignoring gammons and backgammons. Figure 5 shows two runs from this experiment. These runs exhibit an interesting behaviour whereby performance initially increases, but then suffers a noticeable drop before recovering to plateau at about 0.8 against the benchmark opponent. About half of all the runs exhibited this behaviour, with the others showing generally monotonic performance increases but still achieving a similar final level of performance.

To ensure that this behaviour was not an artefact of the benchmark opponent, we re-generated the performance curves with other benchmark opponents. These were created by randomly generating function approximators and training them until they were of similar strength[2] to PUBEVAL. The results were unchanged, the individual curves shifting up or down slightly with the variation in the playing strength of the benchmark opponent, but the characteristic policy degradation remained.

Obviously the downward slope of the dip represents a degradation in the policy during some of the training. However, to be certain that it is the same effect as demonstrated in the previous section, we need to show that TD($\lambda$) is making progress in minimising $\|\tilde{J} - J^{\mu}\|_D$ whilst the policy is degrading. This is not as straightforward in the case of a backgammon system using TD($\lambda$) to train by self-play, because TD($\lambda$) does not passively observe the play, but rather regularly modifies the player. Hence the policy $\mu$ changes each time the neural network weights are updated, and so the existing theoretical results no longer apply. To overcome this, we modified our system to make the TD($\lambda$) updates on the parameters of a non-playing second approximator which was initialised as a copy of the one being observed.

[2]This is necessary because if they were significantly weaker or stronger, the detail of the effect we wish to observe would be obscured as the curve is compressed towards the top or bottom axis.

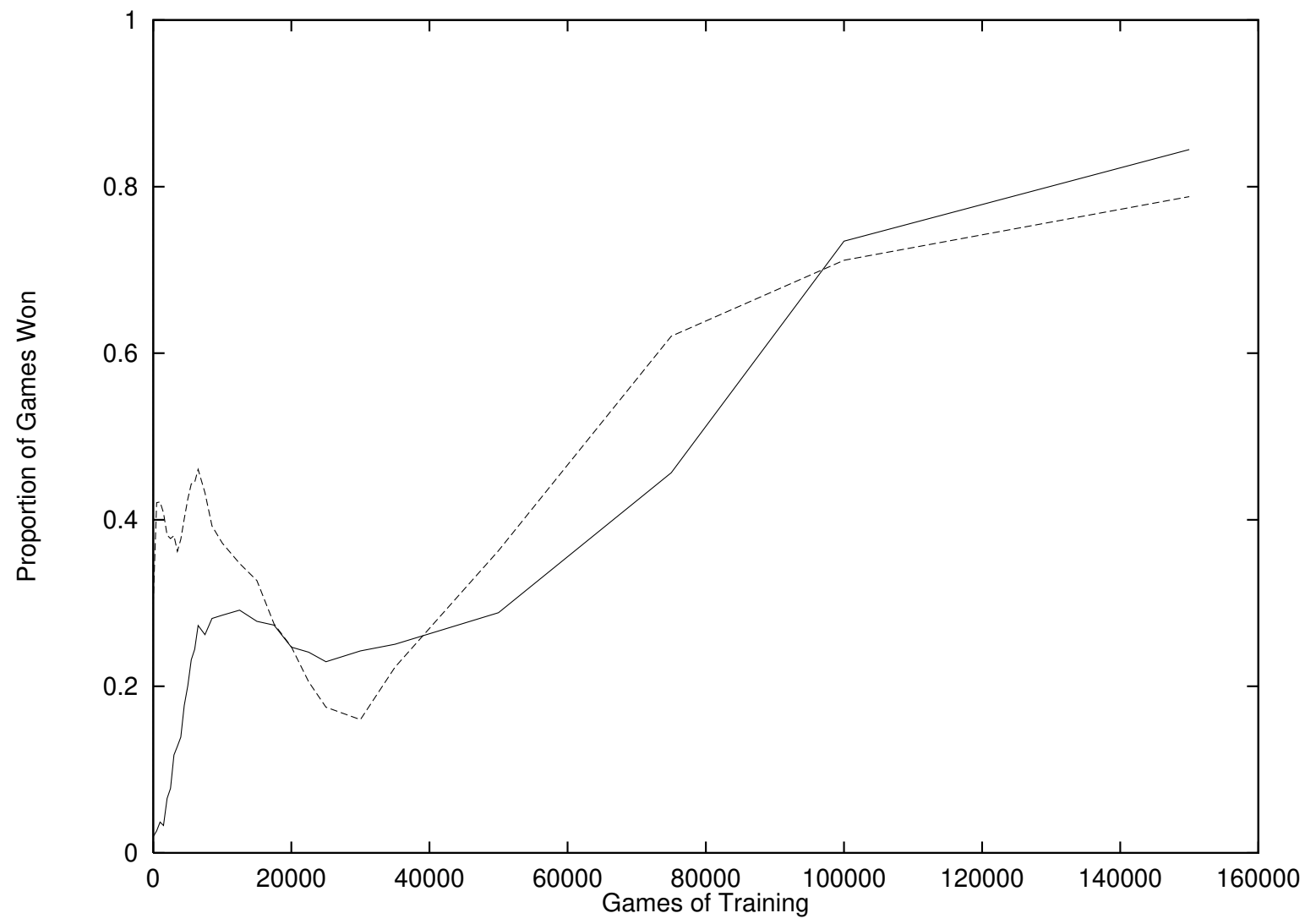

Figure 5: Performance of the function approximator for backgammon, as it trains by self-play with TD($\lambda$). Learning rate $= 0.003$, $\lambda = 0.0$ For smaller learning rates and/or larger $\lambda$, the effect is less pronounced but still present.

As we are interested the policy degradation in Figure 5, we used this modified system to train a network taken from the earlier experiment. The network selected had been trained for 6500 games and was chosen because it was one of the sequence of networks exhibiting policy degradation, i.e. training it under the original TD($\lambda$) regime had resulted in networks implementing inferior policies. We called this network $w_{\text{start}}$.

To estimate $\|\tilde{J} - J^\mu\|_D$ we need to estimate both $J^\mu$ and the distribution $D$. Since the performance of the network is being compared to PUBEVAL, $D$ is the distribution over states generated by play between $w_{\text{start}}$ and PUBEVAL. So to estimate $D$, we collected a large number (2000) positions $S := \{x_1, \dots, x_n\}$ from many games between $w_{\text{start}}$ and PUBEVAL. Each position $x \in S$ was then rolled-out 200 times, and the results averaged to form an estimate $\hat{J}(x)$ of $J^\mu(x)$. Thus $\|\tilde{J} - J^\mu\|_D$ is estimated by $\|\tilde{J} - \hat{J}\|_{\hat{D}}$, where

$$\|\tilde{J} - \hat{J}\|_{\hat{D}} := \frac{1}{n} \sum_{x \in S} \left[ \tilde{J}(x, w) - \hat{J}(x) \right]^2 .$$

Note that $\|\tilde{J} - \hat{J}\|_{\hat{D}}$ is an unbiased estimate of $\|\tilde{J} - J^\mu\|_D$.

Figure 6 shows performance curves (again with PUBEVAL as the opponent) for one run in this experiment. Performance drops from a high of 0.46 to a low of 0.30, a decline of about one-third.

In Figure 7 we see that our estimate of $\|\tilde{J} - J^\mu\|_D$ (called Estimated Error in Approximator), rises initially and then declines throughout the remainder of the run.

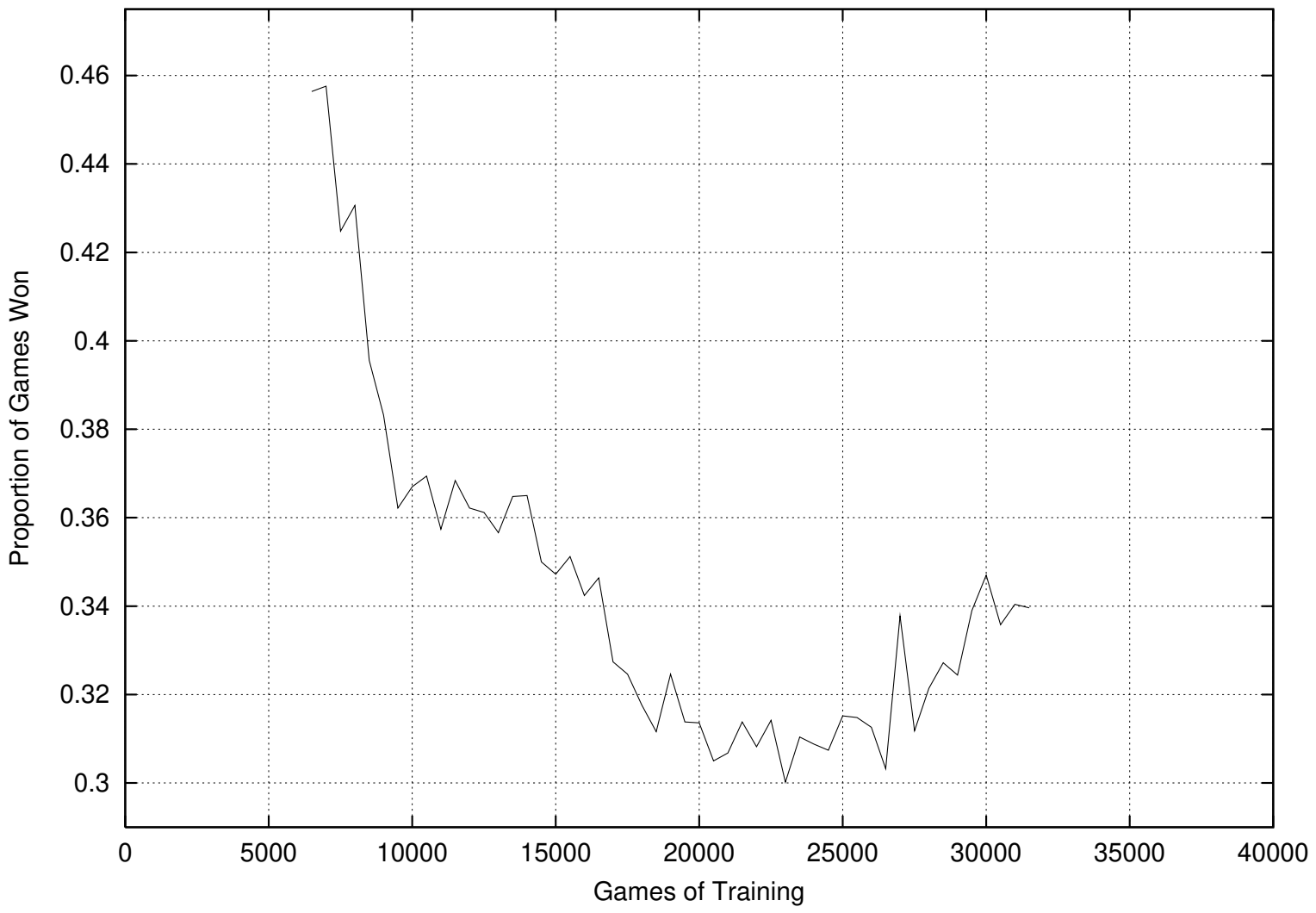

Figure 6: Performance of function approximator for backgammon, as it trains by observation with TD($\lambda$). Learning rate $= 0.003$, $\lambda = 0.0$

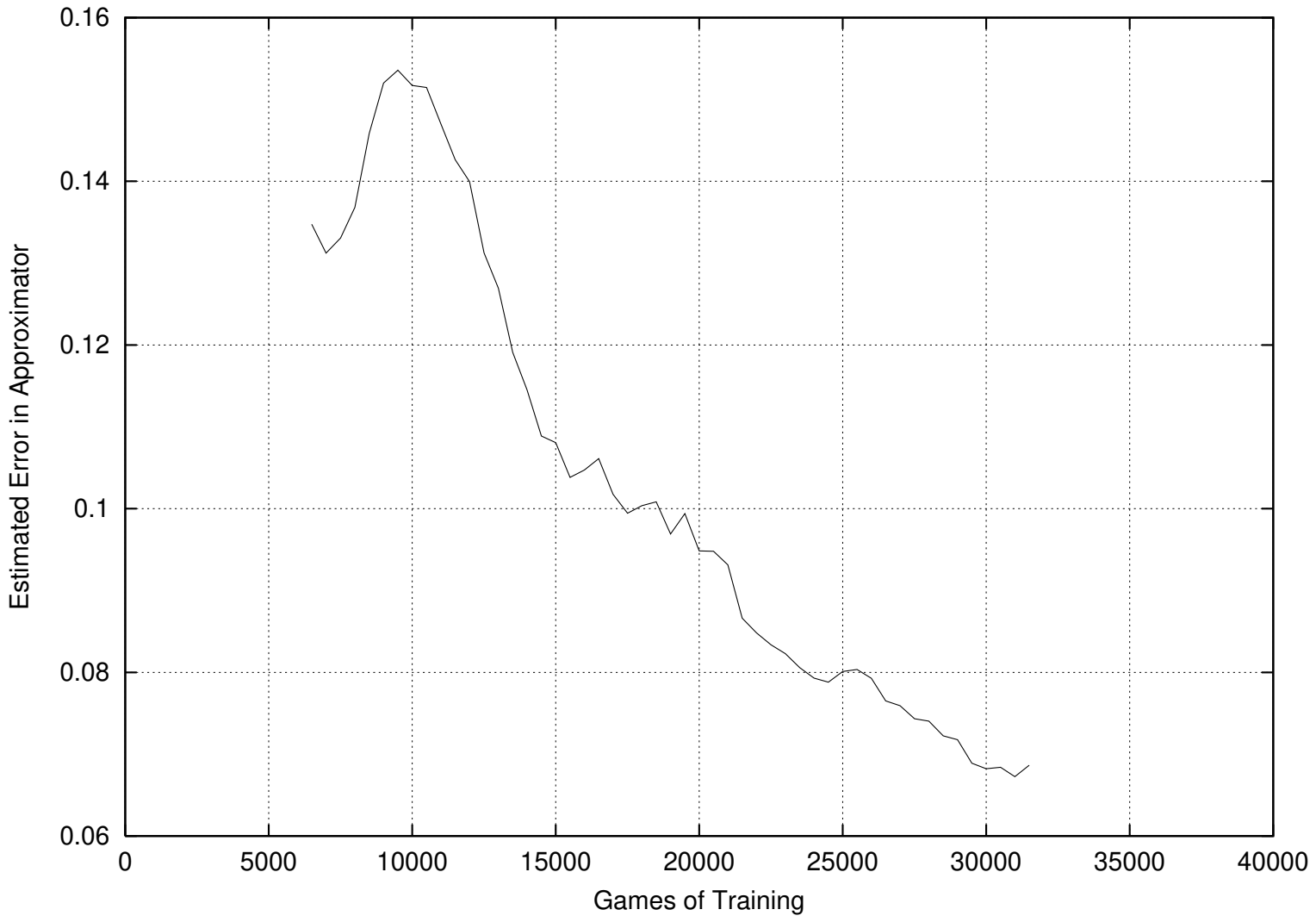

Figure 7: Estimated error in the function approximator for backgammon, as it trains by observation with TD($\lambda$). Learning rate $= 0.003$, $\lambda = 0.0$.

Together these figures are extremely interesting. Normally we would expect to find the error falling with the policy improving concurrently. However, the figures show not this, but two different types of behaviour. Firstly, we see the error rising whilst policy degrades, and secondly we see error falling rapidly as policy continues to degrade.

The first behaviour is not unexpected. If we believe that policy improves as error declines, we should also believe that policy will degrade as error increases. The second behaviour however, clearly demonstrates the policy degradation effect — approximation error falls by half, whilst the performance of the policy continues to decline. Hence, in this case, minimising error in the state-value approximations results in an inferior policy.

# 4 Improving Policy - STD($\lambda$)

The results of the previous two sections raise the question of whether there is a better way of doing TD-style updates. Is there an update rule which precludes policy degradation?

The problem with TD($\lambda$) is that it takes no account of the policy implemented by the function approximator. The updates are not directly derived from the policy, and there is no consideration of the effect on policy that the updates will have. The updates are calculated with the sole intention of moving $\tilde{J}$ closer to $J^\mu$. As we have seen in section 2, the region around $J^\mu$ may implement sub-optimal policies. Even if the region to which we converge has good policies, as in the case of backgammon (section 3), the path which TD($\lambda$) follows may traverse regions of bad policy.

The idea we have tried to capture in STD($\lambda$) is to use the temporal differences to learn relative state values. This endeavours to improve the policy directly, rather than concentrating on approximating individual state values. To simplify matters we have restricted our attention to *Binary Markov Decision Processes* or **BMDP**'s. BMDP's are defined formally in the next section, but loosely speaking, at every state in a BMDP there are at most two states that the system can go to next. We call these states *sibling* states. Unlike TD($\lambda$) which operates on the feature vectors of states directly, STD($\lambda$) operates on the *difference* between feature vectors of sibling states. This is illustrated in Figure 8.

In the remainder of this section we formally define BMDP's and introduce the STD($\lambda$) algorithm. We then characterise the limiting behaviour of STD($\lambda$), in the case of an infinite-horizon, discounted BMDP. Using this result we then prove that in a two-state system with linear function approximation, STD($\lambda$) will never produce a *worse* policy when started from arbitrary initial parameter settings. This is illustrated on the two-state system of section 2.

## 4.1 Binary Markov Decision Processes

A Markov Decision Process (MDP) consists of a set of states $S = 1, \dots, n$ ($n$ possibly countably infinite), a set of actions $A$, a transition probability matrix $P = [p(i, j, a)]$ giving the probability of moving from state $i$ to state $j$ after taking action $a \in A$, and a scalar cost function $g(i, j)$ giving the cost associated with the transition from $i$ to $j$. A policy $\mu$ maps each state to a probability distribution over actions (for simplicity we consider only stationary policies).

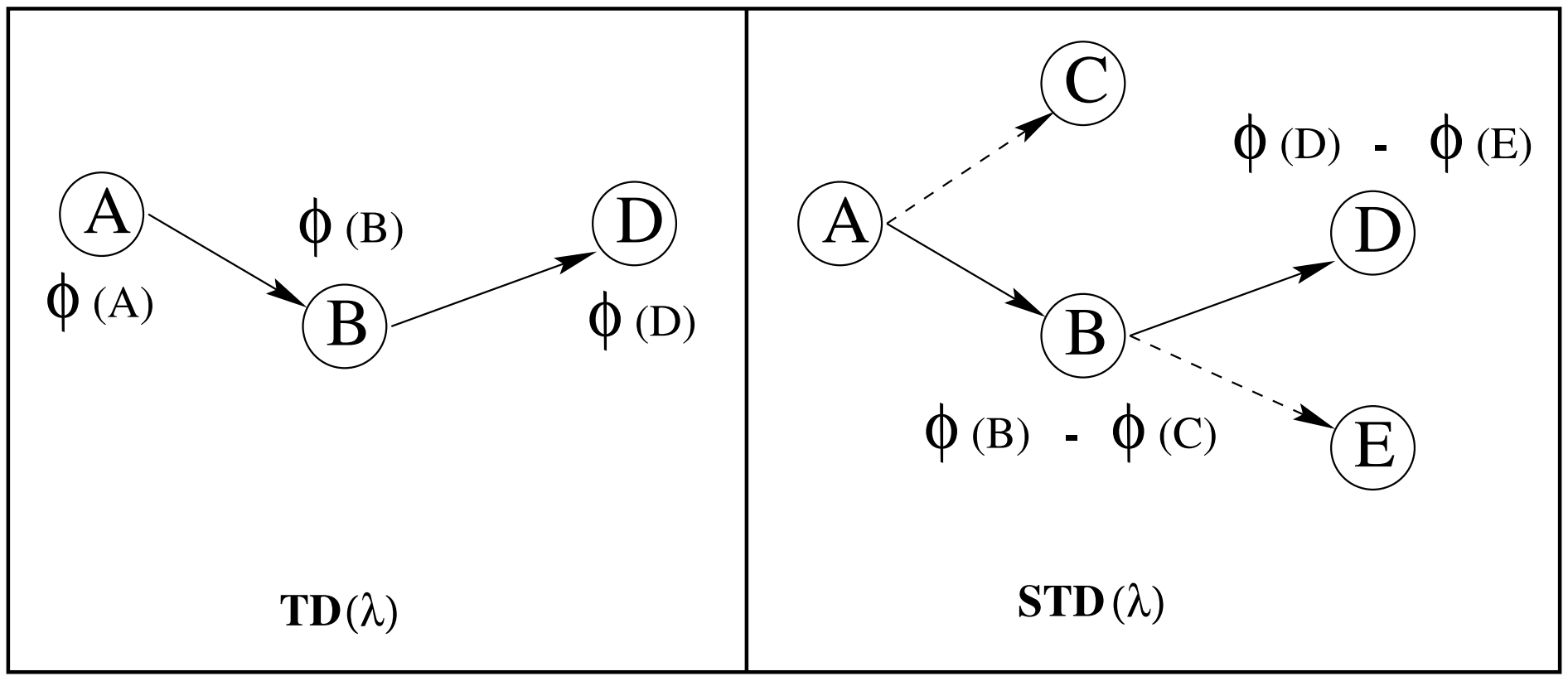

Figure 8: A sequence of states in a Binary Markov Decision Process (BMDP) and how they are viewed by TD($\lambda$) and STD($\lambda$). The solid lines represents the path taken by the process, $A \to B \to D$, with the dotted line representing the alternative paths that might have been followed. TD($\lambda$) only sees the states the system actually visited $(A, B, D)$, and makes updates based on the feature vectors of those states $(\phi(A), \phi(B), \phi(D))$. STD($\lambda$) knows about the alternative paths, or *sibling states*, and makes updates on the *differences* between feature vectors of sibling states.

For any policy $\mu$ and discount factor $\alpha \in [0, 1)$, we define the cost of state $i$ by:

$$J^\mu(i) := \mathbb{E}^\mu \left[ \sum_{t=0}^{\infty} \alpha^t g(x_t, x_{t+1}) | x_0 = i \right],$$

where $x_t$ is the state at time $t$ and $E^\mu$ denotes the expectation over all trajectories under policy $\mu$. An optimal policy $\mu^*$ simultaneously maximizes the reward from every state:

$$J^*(i) = J^{\mu^*} := \max_\mu J^\mu(i).$$

Every MDP has a stationary deterministic optimal policy [21] (here deterministic means each state has an optimal action). Note that under a fixed policy $\mu$ the state transitions $x_t, x_{t+1}, \ldots$ form a Markov chain.

A binary Markov Decision Process (BMDP) is a Markov Decision Process with the property that for each state $i$, $p(i, j, a)$ is non-zero for at most two states $j(i)$ and $j'(i)$. $j(i)$ and $j'(i)$ will be referred to as *sibling states following* $i$ or just *sibling states* if the previous state is clear from the context. In that case we will drop the dependence on $i$ and simply write $j$ and $j'$. If $i$ has only one successor then we set $j' = j$.

## 4.2 The STD($\lambda$) algorithm

We consider a linear approximation to $J^*$ of the form

$$\tilde{J}(i, w) := \sum_{k=1}^{d} w_k \phi_k(i), \tag{4}$$

where $w := (w_1, \ldots, w_k)$ is a vector of tunable parameters and $\phi(i) := (\phi_1(i), \ldots, \phi_k(i))$ is the feature vector associated with state $i$.

The $\mathrm{STD}(\lambda)$ algorithm is described in Algorithm 1. Note that we can obtain a

**Algorithm 1** The $\mathrm{STD}(\lambda)$ algorithm

1: **Given:**

- $\lambda \in [0, 1], \alpha \in [0, 1)$
- State sequence $x_0, x_1, \ldots$ generated by a BMDP under some policy $\mu$.
- Step sizes $\gamma_t > 0$.
- Linear function approximator $\tilde{J}(\cdot, w)$ parameterised by $w \in \mathbb{R}^k$.

2: Choose any starting state $x_0$, initial parameter vector $w_0$, and set $z_0 = 0, (z_0 \in \mathbb{R}^K)$.

3: **for** each state transition $x_t \to x_{t+1}$ **do**

4: $d_t := g(x_t, x_{t+1}) + \alpha \left[\tilde{J}\left(x_{t+1}, w_t\right) - \tilde{J}\left(x'_{t+1}, w_t\right)\right] - \left[\tilde{J}\left(x_t, w_t\right) - \tilde{J}\left(x'_t, w_t\right)\right]$

5: $w_{t+1} := w_t + \gamma_t d_t z_t$

6: $z_{t+1} = \alpha\lambda z_t + \phi(x_{t+1}) - \phi(x'_{t+1})$

7: $t := t + 1$

8: **end for**

version of $\mathrm{STD}(\lambda)$ for nonlinear $\tilde{J}(i, w)$ by replacing step 6 with

$$z_{t+1} := \alpha\lambda z_t + \nabla\tilde{J}(x_{t+1}, w) - \nabla\tilde{J}(x'_{t+1}, w). \tag{5}$$

For our main theorem to hold we need the following assumptions:

**Assumption 1.** *The step sizes $\gamma_t$ are positive and predetermined with $\sum_{t=0}^{\infty} \gamma_t = \infty$ and $\sum_{t=0}^{\infty} \gamma_t^2 < \infty$.*

**Assumption 2.** *The Markov chain generated by the BMDP under policy $\mu$ has a unique stationary distribution $\pi = (\pi(1), \ldots, \pi(n))$.*

**Assumption 3.** *The matrix*

$$\Phi := \begin{matrix} \phi_1(1) & \cdots & \phi_d(1) \\ \vdots & \ddots & \vdots \\ \phi_1(n) & \cdots & \phi_d(n) \end{matrix}$$

*has full rank.*

Since we have assumed the existence and uniqueness of a stationary distribution $\pi(i)$ over states, there exists a stationary distribution $\pi(i, i')$ over states and their siblings. That is, $\pi(i, i')$ is the stationary probability of being in state $i$, with $i'$ as the sibling state. Note that under policy $\mu$,

$$\pi(i, i') = \sum_j \pi(j) p^{\mu}(j, i)$$

where the sum is over all states $j$ with successor states $i$ and $i'$, and $p^{\mu}(j, i)$ is the probability of making a transition from state $j$ to state $i$ under $\mu$.

**Theorem 1.** *Under assumptions 1–3, and with a linear value function* (4)*, the sequence of parameter vectors $w_t$ generated by the* $\mathrm{STD}(\lambda)$ *algorithm converges with probability one. Furthermore, the limiting vector $w_\infty$ satisfies*

$$\sum_{i,j} \pi(i,j) \left[\tilde{J}(i, w_\infty) - \tilde{J}(j, w_\infty) - J^\mu(i)\right]^2 \leq \frac{1-\alpha\lambda}{1-\alpha} \inf_w \sum_{i,j} \pi(i,j) \left[\tilde{J}(i,w) - \tilde{J}(j,w) - J^\mu(i)\right]^2 \quad (6)$$

*Proof.* Under a fixed policy $\mu$ the sequence of pairs of sibling states $(x_t, x'_t), (x_{t+1}, x'_{t+1}), \ldots$ forms a Markov chain with transition probabilities

$$p^\mu\left((i,i'),(j,j')\right) = \begin{cases} p^\mu(i,j) & \text{if } j \text{ and } j' \text{ are siblings following } i, \\ 0 & \text{otherwise} \end{cases}$$

and a stationary distribution $\pi(i,j)$. $\mathrm{STD}(\lambda)$ is simply $\mathrm{TD}(\lambda)$ applied to this derived Markov chain, with an approximate value function given by:

$$\tilde{J}\left((i,i'),w\right) := \tilde{J}(i,w) - \tilde{J}(i',w) = w \cdot [\phi(i) - \phi(i')].$$

Since $\tilde{J}\left((i,i'),w\right)$ is linear in the parameters $w$, the result follows from [9, Theorem 1]. □

Theorem 1 shows that within a factor of $\frac{1-\alpha\lambda}{1-\alpha}$, $\mathrm{STD}(\lambda)$ is aiming for a parameter vector $w$ minimizing the error function:

$$E(w) := \sum_{i,j} \pi(i,j) \left[\tilde{J}(i,w) - \tilde{J}(j,w) - J^\mu(i)\right]^2.$$

Note that each sibling pair $(i,j)$ appears precisely twice in the sum, contributing an amount:

$$E(w,i,j) := \pi(i,j) \left[\tilde{J}(i,w) - \tilde{J}(j,w) - J^\mu(i)\right]^2 + \pi(j,i) \left[\tilde{J}(j,w) - \tilde{J}(i,w) - J^\mu(j)\right]^2,$$

so that

$$E(w) = \frac{1}{2} \sum_{i,j} E(w,i,j).$$

To better understand the solution obtained by minimizing $E(w)$, define

$$\eta(i,j) := \frac{\pi(i,j)}{\pi(i,j) + \pi(j,i)}.$$

Conditioning on the event that we are in a state for which the two successor states are $i$ and $j$, $\eta(i,j)$ is the probability of making a transition to state $i$, whilst $\eta(j,i)$ is the probability of making a transition to state $j$. Note that $\eta(i,j) + \eta(j,i) = 1$. With these definitions, it is easily verified that

$$E(w) = \frac{1}{2} \sum_{i,j} [\pi(i,j) + \pi(j,i)] \left[\tilde{J}(i,w) - \tilde{J}(j,w) - (\eta(i,j) J^\mu(i) - \eta(j,i) J^\mu(j))\right]^2 + K(i,j), \quad (7)$$

where $K(i, j)$ is a function of $J^\mu(i), J^\mu(j), \pi(i, j)$ and $\pi(j, i)$, but does not depend on $w$. Thus, the problem of finding a $w$ minimising $E(w)$ is equivalent to the problem of finding a $w$ minimising $E_1(w)$, where

$$E_1(w) := \sum_{i,j} [\pi(i,j) + \pi(j,i)] \left[\tilde{J}(i,w) - \tilde{J}(j,w) - (\eta(i,j)J^\mu(i) - \eta(j,i)J^\mu(j))\right]^2 . \tag{8}$$

Of course, the quantity we *should* be minimising[3] is:

$$E_2(w) := \sum_{i,j} [\pi(i,j) + \pi(j,i)] \left[\tilde{J}(i,w) - \tilde{J}(j,w) - (J^\mu(i) - J^\mu(j))\right]^2 . \tag{9}$$

However, it seems to be very difficult to minimise (9) directly, principally because for every state, one needs estimates of the value of its sibling states, and these values are not available if we are following a single trajectory (or even a finite number of trajectories). We conjecture that minimising (9) is not possible in general if the memory of the algorithm is constrained.

Comparing (8) and (9), we see that STD is adjusting the approximate differences between sibling states $\tilde{J}(i,w) - \tilde{J}(j,w)$ so that they match the *weighted* true differences $\eta(i,j)J^\mu(i) - \eta(j,i)J^\mu(j)$, rather than the unweighted true differences $J^\mu(i) - J^\mu(j)$. If the weights $\eta(i,j)$ and $\eta(j,i)$ are equal, then STD's target for the sibling pair $(i,j)$ becomes $\frac{1}{2}(J^\mu(i) - J^\mu(j))$, i. e $1/2$ of the true difference[4]. The condition $\eta(i,j) = \eta(j,i)$ corresponds to a policy $\mu$ which has no preference for state $i$ over state $j$. Thus, STD will be optimising the "correct" quantity ($E_2$) whenever it is observing a policy $\mu$ that simply tosses a coin when faced with any decision.

Even if the policy $\mu$ does prefer, say, state $i$ over its sibling $j$, provided

$$\frac{\eta(i,j)}{\eta(j,i)} J^\mu(i) > J^\mu(j), \tag{10}$$

the target $\eta(i,j)J^\mu(i) - \eta(j,i)J^\mu(j)$ will still have the correct *sign*. Note that (10) is always satisfied if the policy $\mu$ is *correct*: i.e $\mu$ prefers $i$ over $j$ and $J^\mu(i) > J^\mu(j)$.

The preceding discussion establishes the following proposition:

**Proposition 1.** *For the two-state model of section 2, let* $\mathrm{STD}(1)$ *be observing a one-step greedy lookahead policy based on the value function* $\tilde{J}(\cdot, w_0)$ *for some initial parameter* $w_0$. *Then the limiting parameter* $w_\infty$ *of* $\mathrm{STD}(\lambda)$ *will itself implement a one-step greedy lookahead policy that is* no worse *than the policy generated by* $\tilde{J}(\cdot, w_0)$.

The same cannot be said of $\mathrm{TD}(1)$ (recall the discussion in section 2).

## 4.3 Comparison with Differential Training

As mentioned earlier, the method of *differential training* described by Bertsekas [1] is the most closely related to the $\mathrm{STD}(\lambda)$ algorithm presented here. During training, it requires two instances of the system being observed to run in lock-step but evolve independently. At time $t$, the state of one of the instances (chosen arbitrarily) is referred to as $x_t$, with the state of the other instance being $\hat{x}_t$. The method uses a TD($\lambda$) approach to learn an approximation, $\tilde{G}(x, \hat{x}, w)$, to the true differences in the values of these states:

$$G^\mu(x_t, \hat{x}_t) = J^\mu(x_t) - J^\mu(\hat{x}_t).$$

[3] In fact, the correct quantity to be optimising is the *expected reward* given parameters $w$. However this is notoriously difficult to do, and is what leads us to consider approximate value-function solutions in the first place. Hence, given that we are working within a squared-loss, *differential* value-function framework, (9) is in some sense the "right" quantity to be optimising.

[4] The factor of $1/2$ is irrelevant in that it just corresponds to a fixed rescaling of all the rewards. It can be removed if we replace the temporal difference $d_t$ in STD with $d_t := 2g(x_t, x_{t+1}) + \alpha\left(\tilde{J}(x_{t+1}, w_t) - \tilde{J}(x'_{t+1}, w_t)\right) - \left(\tilde{J}(x_t, w_t) - \tilde{J}(x'_t, w_t)\right)$.

$G^\mu$ is viewed as the cost-to-go function of a problem involving the compound states $(x, \hat{x})$, and the cost per stage:

$$g(x_t, x_{t+1}) - g(\hat{x}_t, \hat{x}_{t+1}),$$

hence $G^\mu$ satisfies the Bellman equation:

$$G^\mu(x_t, \hat{x}_t) = \mathbb{E}\{g(x_t, x_{t+1}) - g(\hat{x}_t, \hat{x}_{t+1}) + \alpha G^\mu(x_{t+1}, \hat{x}_{t+1})\}.$$

We will refer to the differential training algorithm that uses TD($\lambda$) updates in an infinite horizon discounted reward setting as DT($\lambda$).

Assuming $\tilde{G}$ is linear, and under Assumptions 1–3, DT($\lambda$) will converge (within a factor of $\frac{1-\alpha\lambda}{1-\alpha}$) to a $w$ which minimises:

$$\begin{aligned} E_{DT}(w) &= \sum_{x,\hat{x}} \pi(x)\pi(\hat{x}) \left[\tilde{G}(x, \hat{x}, w) - G^\mu(x, \hat{x})\right]^2 \\ &= \sum_{x,\hat{x}} \pi(x)\pi(\hat{x}) \left[(\tilde{J}(x, w) - \tilde{J}(\hat{x}, w)) - (J^\mu(x) - J^\mu(\hat{x}))\right]^2 \qquad (11) \end{aligned}$$

(by [9, Theorem 1]).

Comparing (11) with (9), we see that DT($\lambda$) is minimising the same quantity we would like to minimise with STD($\lambda$), except that DT($\lambda$) is working with a different distribution over state pairs. This distribution is not the same as the one seen by STD($\lambda$). In particular, the distribution observed by DT($\lambda$) gives a non-zero probability to a pair $(y, \hat{y})$, even if the two component states can never occur as siblings. If $y$ and $\hat{y}$ are both high probability states, then $\pi(y)\pi(\hat{y})$ will be relatively large and the evolution of $w$ will be influenced towards achieving a correct approximation for a pair that will never occur outside the training environment.

In contrast, STD($\lambda$) observes only a single instance of the system during training, and thus trains with respect to the distribution of state pairs occurring under policy $\mu$. State pairs whose components can never occur as siblings will not be observed, and will not influence the evolution of $w$. However, unlike DT($\lambda$), the error function minimised by STD($\lambda$) (8) is not quite the true error function (9). So there is a tradeoff: the true error function can be minimised with respect to the wrong distribution (DT($\lambda$)), or an approximation of the true error function can be minimised with respect to the correct distribution (STD($\lambda$)).

Note that for many dynamical systems (such as physical systems) the number of sibling state-pairs with non-zero probability is $O(n)$, while the total number of state-pairs (and hence the number with non-zero probability under DT($\lambda$)) is always $O(n^2)$. In these systems nearly all the state-pairs considered in DT($\lambda$) are misleading.

## 4.4 Policy degradation in DT($\lambda$)

In this section we present an example 3-state system in which DT($\lambda$) converges to a suboptimal policy while STD($\lambda$) converges to the optimal policy. This example is designed simply to illustrate the pitfalls associated with training against the wrong distribution. In particular, we do not claim STD($\lambda$) always dominates DT($\lambda$).

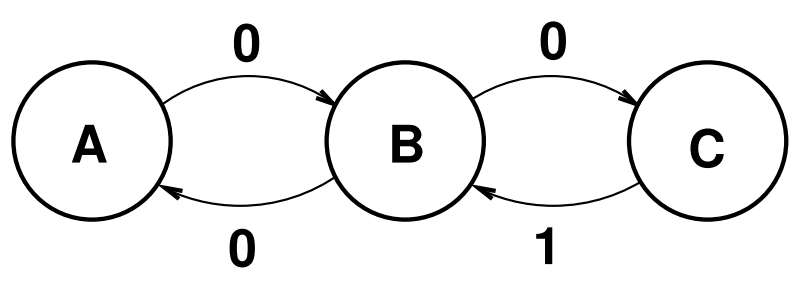

Figure 9: Transitions and rewards in the example three-state system.

The Markov chain with rewards is shown in Figure 9. The only state in which a decision must be made is state $B$, and in that state there are two actions: $a_1$ goes to state $C$ with probability 0.9 and $A$ with probability 0.1, while action $a_2$ goes to state $C$ with probability 0.1 and state $A$ with probability 0.9. Clearly the optimal policy is to always choose action $a_1$.

Under the optimal policy, and for a given discount factor $\alpha \in [0, 1)$, we have $\pi(A) = 0.05, \pi(B) = 0.5, \pi(C) = 0.45$ and

$$J(A) = \frac{0.9\alpha^2}{1-\alpha^2}$$
$$J(B) = \frac{0.9\alpha}{1-\alpha^2}$$
$$J(C) = 1 + \frac{0.9\alpha^2}{1-\alpha^2}.$$

If we take our function approximator to be one-dimensional so that $J(x, w) = w\phi(x)$, and set $\phi(A) = 1, \phi(B) = 3$ and $\phi(C) = 2$, then any positive weight will value state $C$ above state $A$ and hence will implement the optimal policy, while any negative weight will implement the sub-optimal policy. A quick calculation shows that the limiting weight $w_\infty^{\mathrm{DT}}$ of the $\mathrm{DT}(1)$ algorithm satisfies

$$w_\infty^{\mathrm{DT}} = \frac{-0.405 + 0.09\alpha}{0.695}$$

which is negative for all $\alpha \in [0, 1)$, i.e. the wrong sign, while the limiting weight for $\mathrm{STD}(1)$ satisfies

$$w_\infty^{\mathrm{STD}} = 0.9 + \frac{0.72\alpha^2}{1-\alpha^2},$$

which is the right sign for all $\alpha \in [0, 1)$.

# 5 Experimental Results

To illustrate STD($\lambda$) we applied it to the two-state system of section 2. The behaviour of the approximate value function is shown in Figure 10. The system was started with $w = 0.88$ ($\tilde{J}(A, w) = 1.76, \tilde{J}(B, w) = 0.88$; the convergence point of TD($\lambda$) and a sub-optimal policy), it subsequently crossed into the optimal policy region, with $w$ approaching $-0.98$ ($\tilde{J}(A, w) = -1.96, \tilde{J}(B, w) = -0.98$). The same result is achieved for every initial value of $w$.

In a second experiment, we applied both TD($\lambda$) and STD($\lambda$) to the much studied acrobot problem [22, 23, 24]. This problem is analogous to a gymnast swinging on a high bar, and involves simulating a two-link underactuated robot. Torque can be applied only at the second joint.

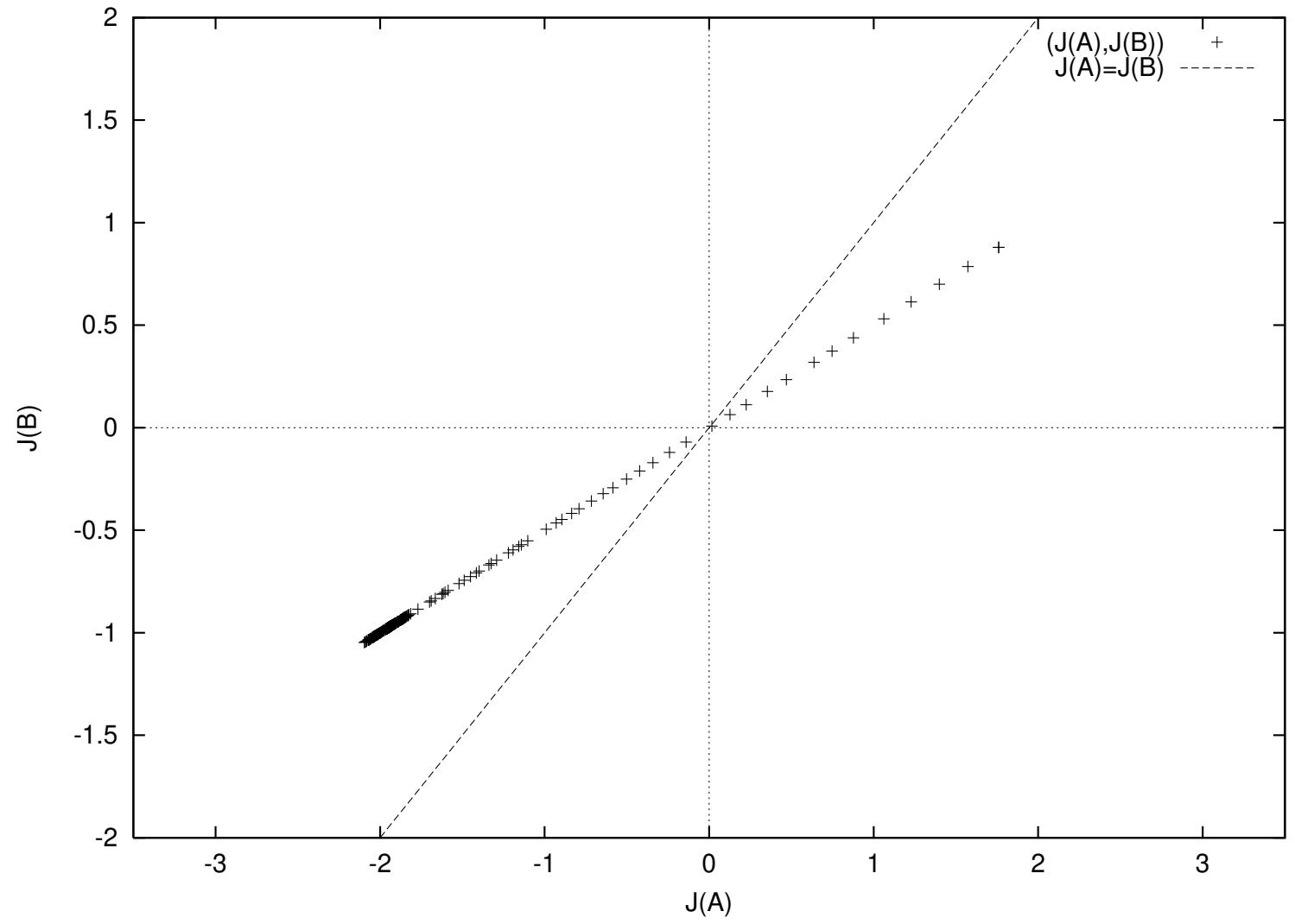

Figure 10: Evolution of $(\tilde{J}(A, w), \tilde{J}(B, w))$ using STD($\lambda$) and starting from $w = 0.88$, ie:$(1.76, 0.88)$, plotted every 50 iterations.

Our implementation is based upon the equations of motion and constants given in [24, page 271], see Figure 11. However, angular velocities are restricted to the interval $[-4\pi, 4\pi]$, and both $\ddot{\theta}_1$ and $\ddot{\theta}_2$ are modified by adding in the damping term $-\frac{|\dot{\theta}_x|}{\dot{\theta}_x} k \dot{\theta}_x^2$ where $k$ is a constant and $x$ is 1 or 2 as appropriate.

For our experiment, the learning algorithms observed a controller which used one-step look-ahead to choose between two actions (torques of +1 and -1 respectively) based on the following evaluation function.

$$J(\theta_1, \theta_2, \dot{\theta}_1, \dot{\theta}_2) = |\dot{\theta}_1 + \dot{\theta}_2| \qquad (12)$$

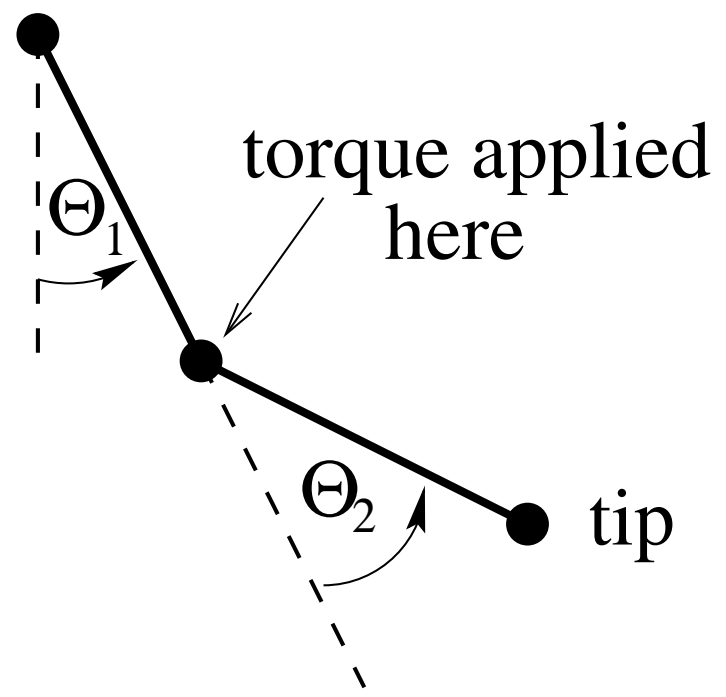

Figure 11: The Acrobot

This function was chosen because it implements a reasonably good policy which is superior to its converse. Using a discount factor of $\alpha = 0.95$, we have empirically estimated the total discounted reward for the policy to be 24.4 (the maximum possible is 80), with an estimate for the converse being 0.4. The nature of our function approximator and feature vector (described below) limit the learning algorithms to choosing between the observed policy and its converse.

Actions were chosen every 0.1 simulated seconds, though motion was simulated at a much finer granularity. The system was run continuously with reward given after simulating the effect of each action choice; the reward being simply the height of the acrobot's tip above its lowest possible position[5].

We used a linear function approximator, and a single element feature vector $\phi(\cdot)$ (see (13) below) which is restricted to the range $[0, 1]$ and gives high values to states *not preferred* by the hand-coded policy based on $J(\cdot)$ (see (12) above).

$$\phi(\theta_1, \theta_2, \dot{\theta}_1, \dot{\theta}_2) = 1 - \frac{|\dot{\theta}_1 + \dot{\theta}_2|}{8\pi} \tag{13}$$

With the acrobot starting from rest in the vertical hanging position, and following the hand-coded policy mentioned above, the two learning algorithms were used to train separate linear evaluators. In both cases $\lambda$ was set to 1.0.

TD($\lambda$) converges to a weight of $w = 13.4$ whilst STD($\lambda$) converges to $w = -394.1$. Since $\phi(\cdot)$ orders states in the opposite order to $J(\cdot)$, a positive value for $w$ means that TD($\lambda$) has converged to the converse of the hand-coded policy, ie: the inferior policy. STD($\lambda$) however, has converged to a negative $w$ and thus its function approximator reverses the ordering imposed by $\phi(\cdot)$ and implements the superior policy.

# 6 Conclusion

We have shown that TD($\lambda$) updates which seek to improve the state value approximation (by minimising $\|\cdot\|_D$) can lead to inferior policies. For the systems detailed in section 2, TD($\lambda$) causes the function approximator to abandon an optimal policy. In section 3 we saw that in a real application, the TD($\lambda$) updates can take the function approximator into regions of parameter space with policies inferior to those that had already been achieved.

For Binary Markov Decision Processes we presented a new algorithm, STD($\lambda$), that retains the advantages of TD($\lambda$) in terms of on-line operation and small memory requirements (only the eligibility trace and current parameter vector need to be stored), but operates directly on the difference in values between *sibling states*, rather than the state values themselves. The limiting behaviour of STD($\lambda$) was characterised for linear function approximators, yielding an interpretation that STD($\lambda$) acts to improve policies rather than the state values themselves. A comparison was made with Bertsekas' DT($\lambda$), and a three-state example presented in which DT($\lambda$) is dominated by STD($\lambda$).

We have also demonstrated that STD($\lambda$) converges to optimal policies on instances of the two-state and acrobot problems where TD($\lambda$) finds only sub-optimal policies.

[5]Both links are 1 metre in length, so the reward for each step is between 0 and 4.